# Generative Nowcasting of Marine Fog Visibility in the Grand Banks area and Sable Island in Canada

Eren Gultepe[1], Sen Wang[2], Byron Blomquist[3], Harindra J.S. Fernando[2],
O. Patrick Kreidl[4], David J. Delene[5], Ismail Gultepe[6]

[1]Southern Illinois University Edwardsville, Edwardsville, IL, USA
`egultep@siue.edu`
[2]University of Notre Dame, South Bend, IN, USA
`swang18, fernando.10 @nd.edu`
[3]NOAA, Boulder, CO, USA
`byron.blomquist@noaa.gov`
[4]University of North Florida, Jacksonville, FL, USA
`patrick.kreidl@unf.edu`
[5]University of North Dakota, Grand Forks, ND
`delene@aero.und.edu`
[6]Ontario Tech University, Oshawa, ON, Canada
`ismail.gultepe@ontariotechu.net`

## Abstract

This study presents the application of generative deep learning techniques to evaluate marine fog visibility nowcasting using the FATIMA (Fog and turbulence interactions in the marine atmosphere) campaign observations collected during July 2022 in the North Atlantic in the Grand Banks area and vicinity of Sable Island (SI), northeast of Canada. The measurements were collected using the Vaisala Forward Scatter Sensor model FD70 and Weather Transmitter model WXT50, and Gill R3A ultrasonic anemometer mounted on the Research Vessel Atlantic Condor. To perform nowcasting, the time series of fog visibility (Vis), wind speed, dew point depression, and relative humidity with respect to water were preprocessed to have lagged time step features. Generative nowcasting of Vis time series for lead times of 30 and 60 minutes were performed using conditional generative adversarial networks (cGAN) regression at visibility thresholds of Vis < 1 km and < 10 km. Extreme gradient boosting (XGBoost) was used as a baseline method for comparison against cGAN. At the 30 min lead time, Vis was best predicted with cGAN at Vis < 1 km (RMSE = 0.151 km) and with XGBoost at Vis < 10 km (RMSE = 2.821 km). At the 60 min lead time, Vis was best predicted with XGBoost at Vis < 1 km (RMSE = 0.167 km) and Vis < 10 km (RMSE = 3.508 km), but the cGAN RMSE was similar to XGBoost. Despite nowcasting Vis at 30 min being quite difficult, the ability of the cGAN model to track the variation in Vis at 1 km suggests that there is potential for generative analysis of marine fog visibility using observational meteorological parameters.

## 1  Introduction

Marine fog plays a role in a variety of aspects such as aviation (Gultepe et al., 2021a), marine shipping (Fernando et al., 2021), and vegetation (Schemenauer et al., 2016). Due to complex microphysical and dynamic interactions over a range of time and space scales, marine fog can form abruptly in marine environments. The intensity at which marine fog is formed is expressed as visibility (Vis). A previous field project (C-FOG) focused on coastal fog and its life cycle along the coastal areas near Nova Scotia, Canada (Fernando et al., 2021; Gultepe et al., 2021a). Owning to the high variability of the fog life cycle with respect to its formation, development, and dissipation across different domains, the marine fog life cycle is significantly different from coastal fog (Gultepe et al., 2016; Koracin and Dorman, 2017) and requires a separate analysis of Vis.

Marine fog nowcasting using Numerical Weather Prediction (NWP) models may have large uncertainties for time scales less than 6 hours (Gultepe et al., 2017), due to limited knowledge on nucleation processes and microphysical schemes (Gultepe et al., 2017; Seiki et al., 2015). As such, many nowcasting studies are limited because of their large time scale predictions. Predictions using ERA5 (fifth generation European Centre for Medium-Range Weather Forecasts (ECMWF) Atmospheric Reanalysis) do not cover shorter time (< 1 hr) (Shan et al. 2019) and space scales (<1 km) (Pavolonis et al., 2005; Mecikalski et al., 2007). Nowcasting using satellite observations are also limited because infrared (IR) channels have a resolution of usually 1 km and short-wave infrared (SWIR) channels cannot be used accurately for monitoring daytime fog (Gultepe et al., 2021a; Pavolonis et al., 2005).

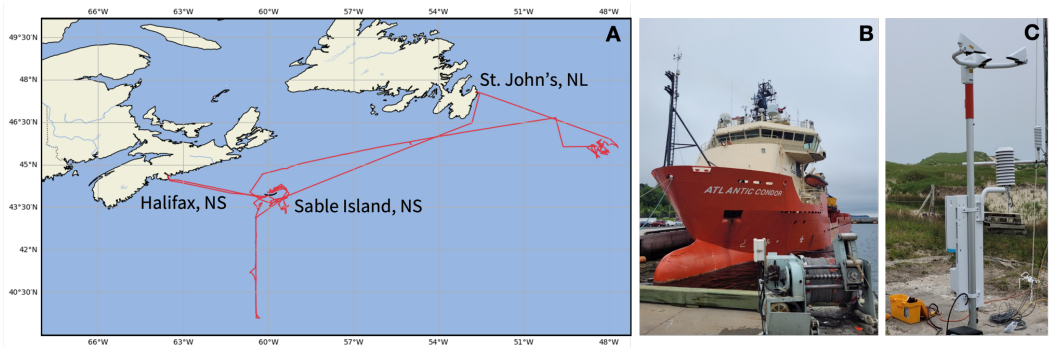

Figure 1: FATIMA field campaign. (a) Campaign mission path. (b) R/V Atlantic Condor with the WXT50 and FD70p instruments located on the upper platform. (c) Close up the FD70p which was used to measure fog visibility.

The limitations of microphysical parameterizations for NWP models used for Vis prediction suggests that alternate techniques are needed for time scales less than 6 hours. Shan et al. (2019) demonstrated that machine learning (ML) techniques can be used for fog Vis prediction. Typically, these studies use ERA5 reanalysis or similar archival datasets, which use hourly, 3-hourly, or daily observations (Pelaez-Rodriguez et al., 2023, Shan et al., 2019). However, fog can form and dissipate quickly in less than a few minutes (Pagowski et al., 2004), thus slow-scale model outputs may not have sufficient resolution to predict faster-scale events. There have been many recent studies that perform binary classification of fog presence (Kamangir et al., 2021; Min et al., 2022; Park et al., 2022; Vorndran et al., 2022). Since such studies mainly use ML classifiers as a blackbox and not for predicting the time series of fog intensity, the dissipation and formation of fog cannot be modelled. Also, these studies analyzed data with a time resolution of 1 hour, obviating the ability to even classify fast dissipating fog.

Very few have studies have pursued predicting the fog visibility time series. Yu et al. (2021) have performed Vis forecasting but the time resolution was limited to 1 hour and for very long lead times of 24 and 48 hours. Also, in the prediction interval there was only one fog event, which provides limited information regarding the variability of Vis. Recent studies have predicted the times series of visibility, but due to pollution rather than fog (Ding et al., 2022; Kim et al., 2022b, 2022a; Zhu et al., 2017) and at a coarse time resolution with very few low visibility events (Yu et al., 2021; Zhu et al., 2017). Although purely deep learning models that inherently account for the autoregressive time series structure have been popular in recent years for various applications such as precipitation nowcasting (Shi et al., 2017; Wang et al., 2017), electricity and traffic load forecasting (Rangapuram et al., 2018), or ride-sharing load forecasting (Laptev et al., 2017; Zhu and Laptev, 2017), such models do not necessarily perform better than non-deep learning techniques (Elsayed et al., 2021; Laptev et al., 2017) unless careful hyperparameter tuning is employed (Laptev et al., 2017). Recent studies that implement generative deep learning models for weather prediction have been used to nowcast precipitation from satellites (Duncan et al., 2022) and calibrate of large-scale neural network weather models (Graubner et al., 2022; Pathak et al., 2022).

For the current study, we used Fog and Turbulence Interactions in the Marine Atmosphere (FATIMA) marine fog field campaign's per-minute observations obtained from various platforms and instruments mounted on the Research Vessel (R/V) Atlantic Condor (Figure 1) collected over the northeast (NE) of Canada, near Sable Island (SI) and Grand Bank (GB) region of the northwest (NW) Atlantic Ocean (Fernando et al. 2022; Gultepe et al. 2023). With these observations, we nowcast marine fog Vis time series by performing regression using conditional generative adversarial networks (cGAN) underpinned by multilayer perceptrons (MLP) (Aggaarwal et al., 2019; Mirza and Osindero, 2014; Oskarsson, 2020). The results demonstrate the possibility of generative modelling of Vis that could potentially complement the extensive field projects such as FATIMA and C-FOG, and possibly help to plan for future campaigns.

## 2  Materials and Methods

The analyzed measurements for the nowcasting of fog Vis (km) were wind speed (Uh in m s$^{-1}$), relative humidity with respect to water (RH$_w$ in %), and dew point depression (Ta-Td in °C), which were all chosen using correlation-based feature selection. The dataset was filtered to only contain observations with precipitation rate (PR) of less than 0.05 mm hr$^{-1}$, since rain visibility can be as large as fog visibility (Gultepe and Milbrandt, 2010). The data was also thresholded to only have observations with, i) Vis < 10 km, to assess how the nowcasting would perform with a variety of high and low visibility; and ii) Vis < 1 km, to assess how nowcasting would perform in a mostly foggy setting (i.e., fog is defined as Vis < 1 km). For all variables, the time-resolution of 1 minute was preserved to mitigate any induced smoothing and aggregation. To perform nowcasting, lagged versions of the input and output variables were created and used in the cGAN regression model (Aggarwal et al., 2019) to directly predict the desired future time step (Hyndman and Athanasopoulos, 2018). In this study, to predict Vis at the 30 min and 60 min lead times, 120 minutes of lag was used for each of Uh, RH$_w$, Ta-Td, and Vis.

To use cGANs for regression (Figure 2), the model architecture is adapted such that the generator receives the meteorological measurements $x$, and noise $z$ (with dimension of 1) and generates the estimate $\hat{y}$ for Vis (Aggarwal et al., 2019). The intuition behind the cGAN architecture is that noise $z$ will capture the latent model error. This cGAN architecture uses the original loss function defined in Mirza and Osindero (2014). The generator and discriminator in the cGAN model use a six-layered and four-layered MLP, respectively and used exponential linear units as the activation function. The learning rates were 0.0001 and 0.001 for the generator and discriminator, respectively. Each network contained 15 neurons per layer, and was trained with a batch size of 100 and ADAM optimizer (Aggarwal et al., 2019). To provide a baseline regression model comparison, boosting with XGBoost (Chen et al., 2015) was used and is commonly used in many fog prediction studies (Vorndran et al. 2022).

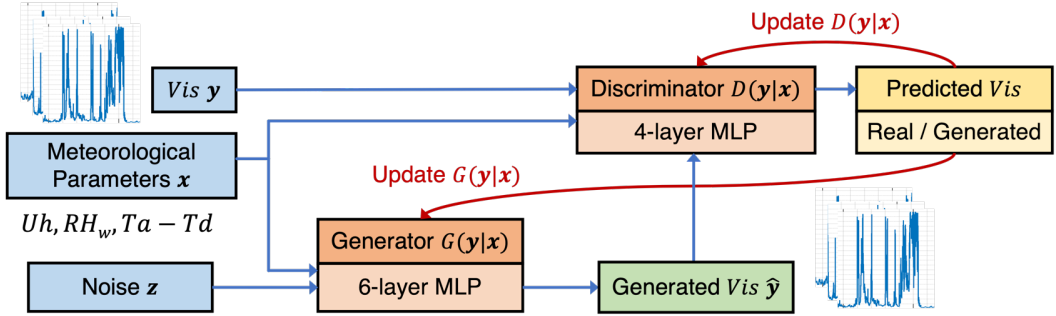

Figure 2: Setup of cGAN for regression to perform nowcasting of fog visibility.

## 3  Results and Conclusions

This is the first study to perform generative nowcasting of marine fog Vis using a cGAN trained with observations that have high temporal resolution and a significant number of fog events. Here we demonstrate the potential for generating fog visibility time series from meteorological parameters, particularly at the 30 min lead time and when fog is present (Vis < 1 km). At the 30 min lead time, Figure 3 and Table 1 show that cGAN (RMSE: 0.151 km) nowcasting had ~20 m better performance than XGBoost (RMSE: 0.170 km) when Vis < 1 km and had only ~15% error in Vis prediction, but at Vis < 10 km their performance differences were negligible. However at the 60 min lead time, Figure 4 shows that XGBoost performed better than the cGAN at both Vis thresholds.

When the performance of the cGAN at Vis 1 < km and 30 min is compared to the naïve forecasting technique of persistence (Per) in which the future prediction is simply the value at the current timestep (Hyndman and Athanasopoulos, 2018), cGAN performed 8.2% better. Whereas the XGBoost had performed 3.3% worse. This is similar to the results of Dietz et al. (2019), who had shown a 5% improvement over Per using boosting at a 30 min lead time. Importantly, at the ≤ 400 m threshold, which is the visibility level that dense fog forms (Yang et al., 2010) and causes severe impediments to maritime activity, the cGAN had a 4.1% improvement over Per.

Compared to the mean of the lookback window (PerW), which is another technique of naïve forecasting (Hyndman and Athanasopoulos, 2018), the cGAN was not as successful, but this is not surprising considering that at 30 min and Vis < 1 km, 74% of the Vis is at the level ≤ 400 m. Thus, predicting the average Vis from the lookback performs well overall, but not does capture the variability of Vis as well as the Per naïve prediction, when the majority of Vis states are very low.

Table 1: RMSE of nowcasting at Vis < 1 km and Vis < 10 km

| Visibility level | Lead time | ML method | Fog Intensity | | |
|---|---|---|---|---|---|
| | | | All | ≤ 400 m | > 400 m |
| Vis < 1 km | 30 min | XGBoost | 0.170 | 0.114 | 0.270 |
| | | **cGAN** | **0.151** | **0.122** | **0.245** |
| | | Per[a] | 0.164 | 0.128 | 0.239 |
| | | PerW[b] | 0.147 | 0.101 | 0.231 |
| | 60 min | XGBoost | 0.167 | 0.110 | 0.268 |
| | | cGAN | 0.209 | 0.170 | 0.288 |
| | | Per | 0.187 | 0.142 | 0.277 |
| | | PerW | 0.156 | 0.106 | 0.248 |
| Vis < 10 km | 30 min | XGBoost | 2.821 | 0.968 | 3.313 |
| | | cGAN | 2.865 | 2.373 | 4.248 |
| | | Per | 2.577 | 1.088 | 2.998 |
| | | PerW | 3.465 | 2.290 | 3.862 |
| | 60 min | XGBoost | 3.435 | 1.691 | 3.948 |
| | | cGAN | 3.508 | 2.697 | 4.200 |
| | | Per | 3.459 | 2.601 | 3.764 |
| | | PerW | 3.928 | 2.717 | 4.340 |

[a]Per, persistence nowcast is calculated as Vis($t+h$) = Vis($t$) where $h$ is either the 30 min or 60 min lead time. [b]PerW, persistence computed using the mean of the Vis in the lookback window of the lagged variables. Bolded RMSE value indicates the best performance of the cGAN nowcasting.

At Vis 10 < km, PerW is a useful comparison because there is more variation in the dense fog versus non-dense fog states and can capture the mean behavior of the Vis time series. Compared to PerW, at 30min and 60min, cGAN performed 17.3% and 10.7% better, respectively. However, compared to Per, cGAN either performed worse (30 min lead time) or approximately the same (60 min lead time). XGBoost had a similar to performance as the cGAN to the Per and PerW naïve predictions. The main difference between the cGAN and XGBoost at Vis < 10 km is that cGAN is not able to track the switching from dense fog to non-dense fog as successfully.

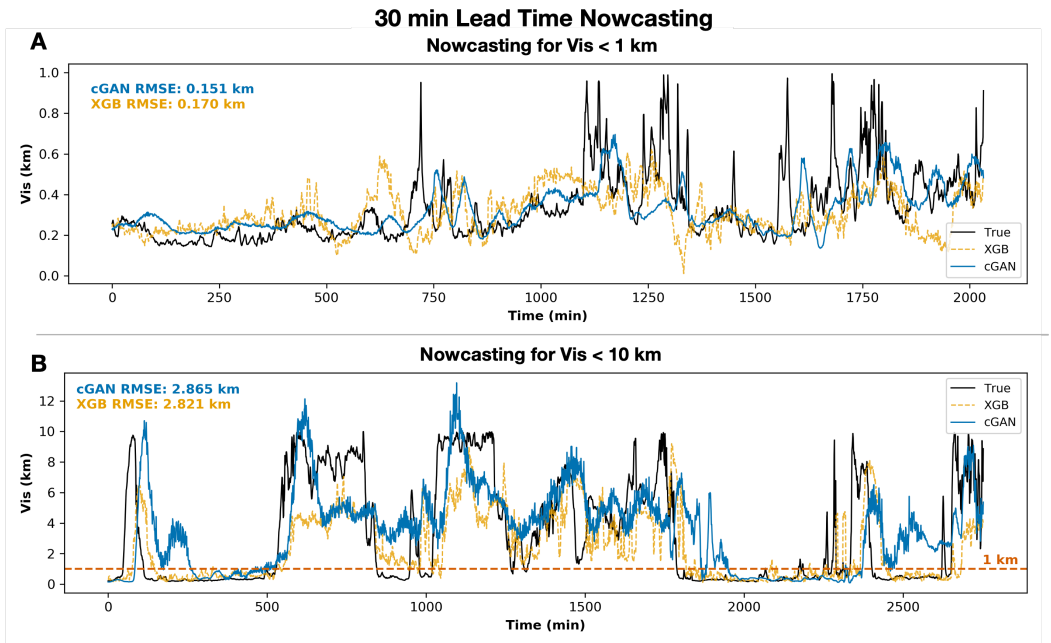

Figure 3: Nowcasting of fog visibility at 30 min lead time. (a) Nowcasting of visibility for intensity of < 1 km. (b) Nowcasting of visibility for intensity < 10 km.

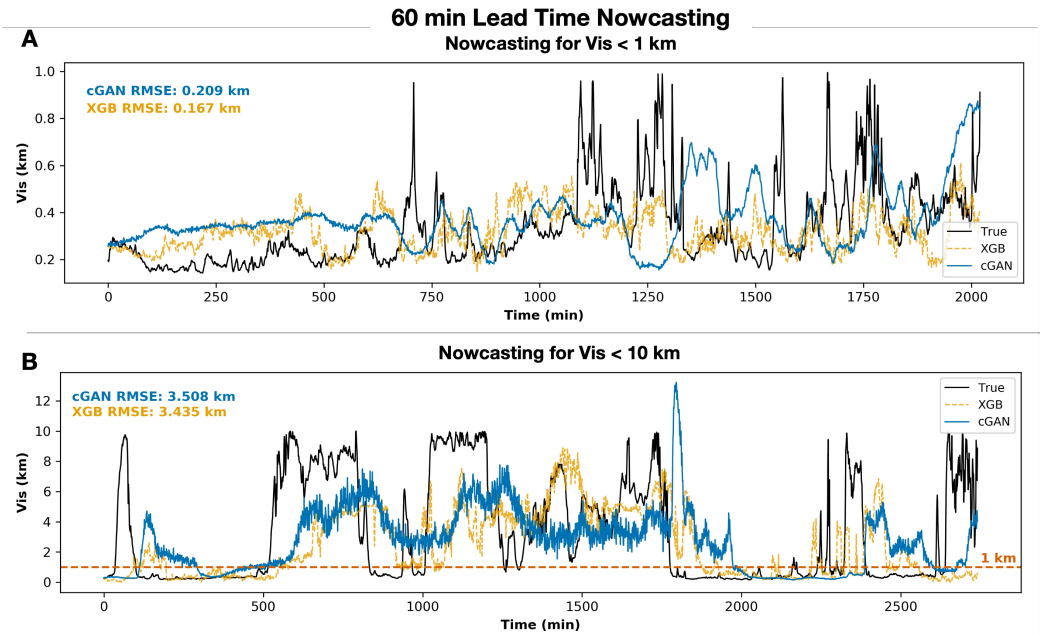

Figure 4: Nowcasting of fog visibility at 60 min lead time. (a) Nowcasting of visibility for intensity of < 1 km. (b) Nowcasting of visibility for intensity < 10 km.

For future studies, we will investigate using RNNs such as LSTMs in place of MLPs in the cGAN models and implement more advanced loss functions that account for the extreme variability in high and low visibility. Also, we would like to focus on even shorter lead times than 30 minutes, as it seems this is when the cGAN had the most success in nowcasting Vis despite the fact that visibility predictions are difficult at short lead times (Dietz et al., 2019, Kneringer et al., 2019).

## Acknowledgments

This project was funded by Grant N00014-21-1-2296 (Fatima Multidisciplinary University Research Initiative) of the Office of Naval Research, administered by the Marine Meteorology and Space Program to Project PI, H. J. S. Fernando.